\pdfoutput=1

\documentclass[11pt]{article}

\usepackage[]{ACL2023}

\usepackage{times}
\usepackage{latexsym}
\usepackage{graphicx}
\usepackage{adjustbox}
\usepackage{booktabs}
\usepackage{multirow}
\usepackage{amsmath}
\usepackage{todonotes}
\usepackage[ruled,vlined]{algorithm2e}
\usepackage[T1]{fontenc}

\usepackage[utf8]{inputenc}

\usepackage{microtype}
\usepackage{amssymb}
\usepackage{pifont}

\usepackage[T1]{fontenc}

\usepackage[utf8]{inputenc}

\usepackage{microtype}

\usepackage{inconsolata}

\title{Semantic-aware Dynamic Retrospective-Prospective Reasoning for Event-level Video Question Answering}

 \author{
  \textbf{Chenyang Lyu}$^\dag$
  ~~~~ \textbf{Tianbo Ji}$^\ddag$~~~~ \textbf{Yvette Graham}$^\P$~~~~ \textbf{Jennifer Foster}$^\dag$~~~~  \\
  $^\dag$ School of Computing, Dublin City University, Dublin, Ireland \\
  $^\ddag$ Nantong University, China \\
  $^\P$ School of Computer Science and Statistics, Trinity College Dublin, Dublin, Ireland\\
  \texttt{chenyang.lyu2@mail.dcu.ie}, \texttt{ygraham@tcd.ie}, \texttt{jennifer.foster@dcu.ie} \\
  \texttt{{jitianbo}@ntu.edu.cn}
}

\begin{document}
\maketitle

\begin{abstract}
Event-Level Video Question Answering~(EVQA) requires complex reasoning across video events to obtain the visual information needed to provide optimal answers. However, despite significant progress in model performance, few studies have focused on using the explicit semantic connections between the question and visual information especially at the event level. 
There is need for using such semantic connections to facilitate complex reasoning across video frames. Therefore, we propose a semantic-aware dynamic retrospective-prospective reasoning approach for video-based question answering. Specifically, we explicitly use the Semantic Role Labeling~(SRL) structure of the question in the dynamic reasoning process where we decide to move to the next frame based on which part of the SRL structure~(agent, verb, patient, etc.) of the question is being focused on. We conduct experiments on a benchmark EVQA dataset - TrafficQA. Results show that our proposed approach achieves superior performance compared to previous state-of-the-art models. Our code will be made publicly available for research use.
\end{abstract}
\section{Introduction}
This paper focuses on one specific variant of Video Question Answering (VQA)~\cite{xu2016msrvtt,yu2018joint-msrvtt-mc,zhong2022videoqa-survey}, namely Event-level VQA~(EVQA)~\cite{xu2021sutd-trafficqa}. In general, the objective of the VQA task is to provide an answer to a visual-related question according to the content of an accompanying video.
 Despite significant recent progress in VQA, EVQA still remains one of the most challenging VQA-based tasks since it requires complex reasoning over the \textit{events} across video frames~\cite{sadhu-etal-2021-video,zhong2022videoqa-survey,liu2022cross-event-level-reasoning-trafficqa}. To tackle the challenges in EVQA, a number of 
approaches have been proposed \cite{xu2021sutd-trafficqa}. \newcite{luo2022temporal-trafficqa} propose a temporal-aware bidirectional attention mechanism for improving event reasoning in videos, while \newcite{zhang2022erm-trafficcqa} propose a novel model named Energy-based Refined-attention Mechanism~(ERM), which obtains better performance compared to previous approaches with a smaller model size. \newcite{liu2022cross-event-level-reasoning-trafficqa}, on the other hand, incorporate visual-linguistic causal dependencies based on Graph Convolutional Networks~\cite{DBLP:conf/iclr/KipfW17-gcn} for enhancing cross-modal event reasoning for EVQA.

\begin{figure*}
    \centering
    \includegraphics[width=0.96\linewidth]{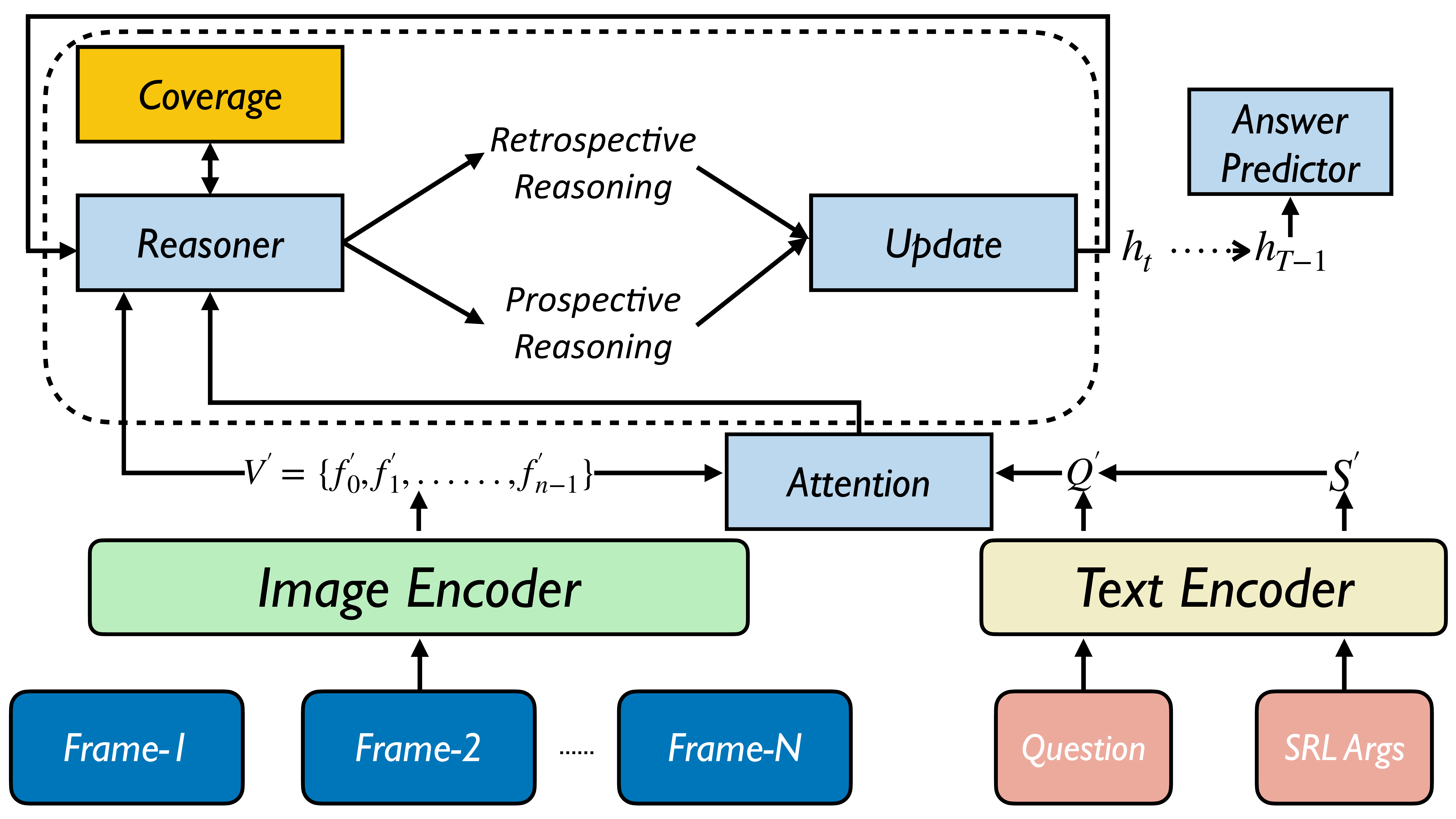}
\caption{Overview of our approach for multi-step visual reasoning. In each reasoning step, the model predicts the reasoning direction (either \textit{retrospective} or \textit{prospective}) and focuses on a specific SRL argument with high attention weights. A \textit{coverage mechanism} is employed to improve the coverage of SRL arguments in the question.}
    \label{fig:model_overview}
\end{figure*}

Despite recent advances, conventional EVQA approaches generally fail to take into account the explicit semantic connection between questions and the corresponding visual information at the event level. Therefore, we propose a new approach that takes advantage of such semantic connections, using the Semantic Role Labeling~(SRL)~\cite{marquez2008semantic-role-labeling,palmer2010semantic-role-labeling,he-etal-2017-deep-srl} structure of questions. The model uses SRL information to learn an explicit semantic connection between the text-based questions and visual information in videos. Additionally, we carry out a multi-step reasoning mechanism over video frames to avoid adapting to spurious correlation and shortcuts by explicitly learning the reasoning process itself~\cite{yi2018neural-reasoning,zhang2021explicit,picco-etal-2021-neural-unification,hamilton2022neuro,zhu2022shallow-compositional-explicit-reasoning}. 

Specifically, in each reasoning step, the model should explicitly decide which frame should be focused on by predicting the reasoning direction~(\textit{retrospective} or \textit{prospective}). In terms of the question, in each reasoning step, we focus on one or more specific SRL arguments with high attention weights, and model its connection with the visual information~(i.e., video frames) contained within the corresponding video. For example, for a question such as \textit{[\textbf{ARG1}: How many cars] were [Verb: involved] [\textbf{ARG2}: in the accident?]}, the model concentrates on the \textit{\textbf{ARG2}} when locating the accident, before determining how many cars were involved in the accident~(\textit{\textbf{ARG1}}). In a specific reasoning step, $t$, we inject the relevant visual information based on the semantic connection between the question and video frames by updating a hidden vector. This vector is ultimately expected to contain the necessary information for predicting the correct answer. In the reasoning process, we employ a \textit{coverage mechanism}~\cite{tu-etal-2016-modeling-coverage} to improve the coverage of the SRL arguments of question. Namely, instead of simply focusing on a small number of specific arguments, the model is capable of including a large range of arguments. 

To investigate the effectiveness  of the proposed approach,
we conduct experiments on a benchmark EVQA dataset: TrafficQA. Results reveal the model to achieve  performance superior to that of existing baselines for a range of reasoning types~(e.g., counterfactual, prospective).

\section{Methodology}
An overview of our approach is shown in Figure~\ref{fig:model_overview}. Suppose the input of our model consists of a video $V$ composed of $n$ image frames sampled from it: $V=\{f_{0}, f_{1},......,f_{n-1}\}$, and a corresponding question $Q=\{w_{0}, w_{1},......,w_{m-1}\}$ with associated SRL arguments $S=\{S_{0}, S_{1},......,S_{N-1}\}$ where $S_{i}=\{w_{i}, w_{i+1},......,w_{k}\}$. All frames $V=\{f_{0}, f_{1},......,f_{n-1}\}$ are fed into an \textsc{Image Encoder} followed by temporal attention modeling to produce temporal-aware frame representations $V^{'} = \{f_{0}^{'}, f_{1}^{'},......,f_{n-1}^{'}\}  \in \mathbf{R}^{n\times d}$. Meanwhile, we use a \textsc{Text Encoder} to obtain the representations of the question with its corresponding SRL arguments: $Q^{'}\in\mathbf{R}^{1\times d}$ and $S^{'}\in\mathbf{R}^{N\times d}$. We then perform multi-step reasoning in which we iteratively update the hidden state vector $h$ with the visual information from frame representations based on the attention weights between them and the SRL arguments of the question. $h$ is updated from the initial step $h_{0}$ to the final step $h_{T-1}$ where $T$ is the total number of reasoning steps. Finally, we predict the most probable answer $a$ based on $h_{T-1}$.
\subsection{Multi-step Reasoning}
Before the first reasoning step, we initialize:
\begin{equation}
    h_{0} = Attn(Q^{'}, V^{'}, V^{'})
\end{equation}
\begin{equation}
    j = argmax(AttnWeights(Q^{'}, V^{'}, V^{'}))
\end{equation}
 where $Attn$ serves as the $q,k,v$ \textit{attention}\footnote{In this work, we use a low temperature $\tau$ in the \textit{softmax} to encourage the model to assign more attention weights to the most relevant frame.} modeling~\cite{transformer} and $j$ represents the index of the frame with the highest attention weight. In each specific reasoning step $t$, we firstly use $h_{t-1}$ as the \textit{attention key} to obtain the relevant SRL argument: $S^{'}_{t} = Attn(h_{t-1}, S^{'}, S^{'})$. Subsequently, we infer the next focused frame by:

\begin{equation}
\label{equation:frame_v_update}
  V^{focus}= Attn(r_{t}, V^{'}, V^{'})
\end{equation}
 
 where $r_{t} = g(h_{t-1}, S_{t}^{'})$. Finally, we update the hidden state vector $h_{t-1}$ based on the currently focused frame~(the frame with the largest attention weight):
\begin{equation}
 \label{equation:hidden_h_update}
    h_{t} = \delta(h_{t-1}, V^{focus})
\end{equation}

\subsection{Retrospective-Prospective Reasoning}
We propose a \textit{Retrospective-Prospective Reasoning} mechanism for Eq.\ref{equation:frame_v_update} in order to explicitly decide whether the model should move to future frames~(\textit{prospective reasoning}) or move back to previous frames~(\textit{retrospective reasoning}). We obtain the \textit{retrospective frame} $V^{retro} $ and \textit{prospective frame} $V^{prosp} $ by:

\begin{equation}
      \small
      V^{retro} = \psi(g(h_{t-1}, S^{'}_{t}), V^{'}, RetroMask(j))
\end{equation}
\begin{equation}
      \small
      V^{prosp} = \phi(g(h_{t-1}, S^{'}_{t}), V^{'}, ProspMask(j))
\end{equation}

where $\psi$ and $\phi$ are \textsc{Masked Attention} that are used to obtain \textit{retrospective} and \textit{prospective} frames, $g(h_{t-1}, S^{'}_{t})$ and $V^{'}$ serve as \textit{query} and \textit{key, value} respectively. $RetroMask(j)$ means all frames after $j$~($f_{i>j}$) will be masked whereas $ProspMask(j)$ means that all frames before $j$~($f_{i<j}$) will be masked. After obtaining $V^{retro} $ and $V^{prosp} $ we generate a probability:
\begin{equation}
    p=\sigma(\lambda(V^{retro} ,V^{prosp} ))
\end{equation}
 
If $p$ is larger than a pre-defined threshold $\alpha$, we update $h_{t}=\delta(h_{t-1}, V^{retro} )$ ,otherwise we update $h_{t}=\delta(h_{t-1}, V^{prosp} )$ as in Eq.~\ref{equation:hidden_h_update}. The index for the next-focused frame $j$ is also updated accordingly. We present further details of our algorithm in the Appendix. 

\subsection{Coverage Mechanism}
We additionally propose to employ a \textit{coverage mechanism}~\cite{tu-etal-2016-modeling-coverage} to encourage the model to include as many SRL arguments as possible in the reasoning process. Specifically, we track the attention distribution $C_{t}\in\mathbf{R}^{1\times N}$ of $h_{t-1}$ on all SRL arguments $S$
\begin{equation}
\small
    C_{t} = C_{t-1} + \frac{AttnWeights([h_{t-1};C_{t-1}], S^{'}, S^{'})}{\chi}
\end{equation}
where $\chi$ represents the normalization factor.\footnote{In this work, we use the number of SRL arguments of the corresponding question as the normalization factor.} We obtain the weighted $S_{t}^{'}$ by $S^{'}_{t} = Attn([h_{t-1};C_{t-1}], S^{'}, S^{'})$ where we concatenate $C_{t-1}$ to $h_{t-1}$ as an additional input to the \textit{Attn} function for the purpose of informing the model to assign more attention weights to previously less-focused SRL arguments, in order to improve the 
coverage for all SRL arguments.

\subsection{Training Objective}

For the answer prediction, we encode all answer options $A=\{a_{0},......,a_{M-1}\}$ separately and then select the one with the highest similarity with $h_{T-1}$. We optimize our model parameters $\theta$ using \textit{Cross Entropy} loss:

\begin{equation}
    J(\theta) = -\sum_{i}\sum_{k}log\frac{e^{\digamma(a_{k},h_{T-1})}}{\sum_{j=0}^{M-1}e^{\digamma(a_{j},h_{T-1})}}y_{i,k}
\end{equation}

where $\digamma$ is the function measuring the similarity between answer candidate and $h_{T-1}$, and $y_{i,k}$ represents the answer label for the $i-$th example - if the correct answer for the $i-$th example is the $k-$th answer then $y_{i,k}$ is 1 otherwise it is 0.

\section{Experiments}
\begin{table}
\resizebox{\linewidth}{!}{
\begin{tabular}{lccc}
\toprule
Models & Setting-1/4 & Setting-1/2   \\ \midrule
Q-type (random)~\cite{xu2021sutd-trafficqa}& 25.00 & 50.00  \\ 
QE-LSTM~\cite{xu2021sutd-trafficqa} & 25.21 & 50.45  \\ 
QA-LSTM~\cite{xu2021sutd-trafficqa} & 26.65 & 51.02   \\ 
Avgpooling~\cite{xu2021sutd-trafficqa} & 30.45& 57.50 \\ 
CNN+LSTM~\cite{xu2021sutd-trafficqa} & 30.78 & 57.64  \\ 
I3D+LSTM~\cite{xu2021sutd-trafficqa} & 33.21 & 54.67  \\
VIS+LSTM~\cite{ren2015exploring} & 29.91 & 54.25\\
BERT-VQA~\cite{Yang_2020_WACV} & 33.68 & 63.50 \\ 
TVQA~\cite{lei2018tvqa} & 35.16 & 63.15   \\ 
HCRN~\cite{Le_2020_CVPR} & 36.49 & 63.79  \\ 
Eclipse~\cite{xu2021sutd-trafficqa}  & {37.05} & {64.77} \\ 
ERM~\cite{zhang2022erm-trafficcqa} & 37.11 & 65.14 \\
TMBC~\cite{luo2022temporal-trafficqa} & 37.17 & 65.14 \\
CMCIR~\cite{liu2022cross-event-level-reasoning-trafficqa} & 38.58 & N/A \\
Ours & \textbf{43.19} & \textbf{71.63} \\
\bottomrule

\end{tabular}
}
\caption{Evaluation results on TrafficQA dataset.}
\label{table:baseline_comparison}
\end{table}

\subsection{Dataset}
We employ a benchmark dataset for EVQA - TrafficQA~\cite{xu2021sutd-trafficqa}  which contains 62,535 QA pairs and 10,080 videos. We follow the standard split of TrafficQA -- 56,460 pairs for training and 6,075 pairs for evaluation. 
We further sample 5,000 examples from training data as the dev set. 

\begin{table*}[t]\renewcommand\tabcolsep{6.0pt}\renewcommand\arraystretch{1}
\begin{center}
\resizebox{\linewidth}{!}{
\begin{tabular}{lccccccc}
\toprule
\multirow{2}*{Method}&\multicolumn{6}{c}{Question Type}\\\cline{2-8}
&Basic&Attribution&Introspection&Counterfactual&Forecasting&Reverse&All\\ \midrule
\textrm{HCRN}~\cite{le2020hierarchical}&34.17&50.29&33.40&40.73&44.58&50.09&36.26\\
\textrm{VQAC}~\cite{kim2021video}&34.02&49.43&34.44&39.74&38.55&49.73&36.00\\
\textrm{MASN}\cite{seo-etal-2021-attend}&33.83&50.86&34.23&41.06&41.57&50.80&36.03\\
\textrm{DualVGR}~\cite{wang2021dualvgr}&33.91&50.57&33.40&41.39&41.57&50.62&36.07\\
CMCIR~\cite{liu2022cross-event-level-reasoning-trafficqa}&36.10  &52.59&38.38  &46.03&48.80&52.21&38.58\\

Ours& \textbf{37.05}  & \textbf{52.68} & \textbf{43.91}  & \textbf{50.81} & \textbf{54.26} & \textbf{55.52} & \textbf{43.19} \\
\bottomrule
\end{tabular}
}
\end{center}
\caption{Results by various \textit{question type} on the dev set of TrafficQA. The highest performance are in bold.}
\label{Table4}
\end{table*}
\subsection{Experimental Setup}
We use CLIP ViT-B/16~\cite{radford2021learning-clip}~\footnote{https://openai.com/blog/clip/} to initialize our image encoder and text encoder. We evenly sample 10 frames from each video in the TrafficQA dataset. The SRL parser employed in the experiments is from AllenNLP~\cite{gardner-etal-2018-allennlp,Shi2019SimpleBM}. We train our model over 10 epochs with a learning rate of $1\times 10^{-6}$ and a batch size of 8. The optimizer is AdamW~\cite{adamw}. We set the maximum reasoning step $T$ to 3 and we use a temperature $\tau$ of 0.2 in \textit{Attention} modeling. The hyper-parameters are empirically selected based on the performance on dev set. There are two experimental settings for TrafficQA~\cite{xu2021sutd-trafficqa}: 1) Setting-1/2, this task is to predict whether an answer is correct for a given question based on videos; 2) Setting-1/4: this task follows the standard setup of multiple-choice task in which the model is expected to predict the correct the answer from the four candidate options.

\subsection{Results}
The experimental results on the 
test set of TrafficQA are shown in Table~\ref{table:baseline_comparison}, where we also include the previous baseline models for EVQA.\footnote{Some of the baseline results are taken from~\newcite{xu2021sutd-trafficqa}.} The results 
show that our proposed approach obtains accuracy of 43.19 under the multiple-choice setting, which surpasses previous state-of-the-art approaches including Eclipse~\cite{xu2021sutd-trafficqa}, ERM~\cite{zhang2022erm-trafficcqa}, TMBC~\cite{luo2022temporal-trafficqa} and CMCIR~\cite{liu2022cross-event-level-reasoning-trafficqa} by at least 4.5 points. Furthermore, our approach achieves an accuracy of 71.63 under Setting 1/2, outperforming previous strong baselines by at least 6 points. The results show the effectiveness of our proposed multi-step reasoning approach for event-level VideoQA.

\paragraph{Ablation Study} We conduct experiments on the dev set of TrafficQA, investigating the contribution of both the \textit{retrospective-prospective reasoning} and \textit{coverage mechanism} on the performance of our proposed EVQA approach. The results are shown in Table~\ref{table:ablation_study}, which reveals that multi-step reasoning is critical in terms of model performance while the \textit{coverage mechanism} can provide additional, albeit less substantial, improvements.

\begin{table}[t]
\setlength{\tabcolsep}{10pt}
\begin{center}
\resizebox{\linewidth}{!}{
\begin{tabular}{lccc}
\toprule
Models & Setting-1/4 & Setting-1/2   \\ 
\midrule
Model w/o MR and CM & 42.53 &  69.61 \\
Model w/o CM & 46.15 & 74.97 \\ 
 Model & 47.38 & 75.83  \\ 
 \bottomrule
\end{tabular}
}
\end{center}
\caption{Ablation study results on TrafficQA dev set, where \textit{MR} represents \textit{Multi-step Reasoning} and \textit{CM} represents \textit{Coverage Mechanism}. MR and CM are coupled in our approach.}
\label{table:ablation_study}
\end{table}

\paragraph{Results by Question Type} We take a closer look at model performance on different question types, e.g. reverse reasoning, counterfactual reasoning, etc. The results are shown in Table~\ref{Table4}. They reveal that our proposed approach outperforms previous state-of-the-art models on all individual question types by a large margin with large improvements seen for \textit{introspection}, \textit{reverse} and \textit{counterfactual} questions.

\paragraph{Effect of Reasoning Steps}
We study the effect of varying reasoning steps. The results are shown in Table~\ref{table:effect_of_reasoning_steps}.  Increasing reasoning steps improves performance, especially from 1 step to 3 steps. Additionally, the performance~(both Setting 1/4 and 1/2) is stable with reasoning steps exceeding three.

\begin{table}[t]
\setlength{\tabcolsep}{10pt}
\begin{center}
\resizebox{\linewidth}{!}{
\begin{tabular}{lccc}
\toprule
Reasoning Steps & Setting-1/4 & Setting-1/2   \\ \midrule
Model w/ 1 step & 41.57 & 71.46 \\ 
Model w/ 2 steps & 44.21 & 74.95  \\ 
Model w/ 3 steps & 47.38 & 75.83  \\ 
Model w/ 4 steps & 47.23 & 75.96  \\ 
Model w/ 5 steps & 47.15 & 75.87 \\  \bottomrule
\end{tabular}
}
\end{center}
\caption{The effect of various reasoning steps.}
\label{table:effect_of_reasoning_steps}
\end{table}

\section{Conclusion and Future Work}

In this paper, we propose a multi-step dynamic retrospective-prospective approach for EVQA.
Our approach employs a multi-step reasoning  model that explicitly learns reasoning based on the semantic connection of the  SRL structure of a question and corresponding video frames. We additionally proposed a \textit{coverage mechanism} to improve the coverage of SRL arguments in the reasoning process. Experimental results show that the proposed approach obtains superior performance compared to that of state-of-the-art EVQA models.

\section*{Limitations}
This papers focuses on a variety of VideoQA - event-level VideoQA, we only incorporate \textit{event} information from the question~(textual) side as we think that parsing video frames is inaccurate and could introduce unexpected errors, we should also explore how to inject \textit{event-level} information from visual side in the future with more competitive visual parsing models. Our experiments are only conducted on one dataset due to resource constraint, we should also conduct experiments on more datasets to verify the effectiveness of our approach.




\bibliography{anthology,custom,trafficqa_arxiv}
\bibliographystyle{acl_natbib}

\appendix

\section{Appendix}
\label{sec:appendix}

\subsection{Algorithm of Multi-step dynamic retrospective-introspective reasoning}
\begin{algorithm}
\small

$ V^{'}=\{f_{0}, f_{1},......,f_{n-1}\}$: {representations of video frames}

$Q^{'}$: {question}

$S^{'}$: {SRL representations of $Q$}

$T$: {reasoning steps}

$\chi$ : {normalization factor}

$\alpha$: {threshold of the probability for using retrospective frame}

$h_{0} = Attn(Q^{'}, V^{'}, V^{'})$

$j = argmax(AttnWeights(Q^{'}, V^{'}, V^{'}))$

$C_0 = 0$

 \For{$i$ in $T$}{

  $S^{'}_{i} = Attn(h_{i-1}, S^{'}, S^{'}, C_{i-1})$

  $C_{i} = C_{i-1} + \frac{AttnWeights(h_{i-1}, S^{'}, S^{'}, C_{i-1})}{\chi}$

  $ V^{retro} = \psi(g(h_{t-1}, S^{'}_{t}), V^{'}, RetroMask(j))$

  $V^{prosp} = \phi(g(h_{i-1}, S^{'}_{i}), V^{'}, ProspMask(j))$

  $p = \sigma(f(V^{retro} , V^{prosp} ))$

  \If{$p > \alpha$}{
        $h_{i} = \delta(h_{i-1}, V^{retro} )$
        
          $j = argmax(\psi(g(h_{t-1}, S^{'}_{t}), V^{'}, RetroMask(j)))$
  }
  \Else{
        $h_{i} = \delta(h_{i-1}, V^{prosp} )$
        
       $j = argmax(\phi(g(h_{i-1}, S^{'}_{i}), V^{'}, ProspMask(j)))$

  }
  
 }

 \caption{Multi-step dynamic retrospective-prospective reasoning with coverage mechanism}
 \label{multi-step-reasoning}
\end{algorithm}


\end{document}